\documentclass[twoside,11pt]{article}

\usepackage{jmlr2e}
\usepackage{algorithmic}
\usepackage{algorithm}
\usepackage{amsmath, amssymb}
\usepackage{multirow}
\usepackage{subcaption}
\usepackage{rotating}
\usepackage{mathtools}
\usepackage{grffile}
\graphicspath{ {images/} }

\ShortHeadings{Soft-Label Dataset Distillation and Text Dataset Distillation}{Sucholutsky and Schonlau}
\firstpageno{1}

\begin{document}

\title{Soft-Label Dataset Distillation and Text Dataset Distillation}

\author{\name Ilia Sucholutsky \email isucholu@uwaterloo.ca \\
       \addr Department of Statistics and Actuarial Science\\
       University of Waterloo\\
       Waterloo, Ontario, Canada
       \AND
       \name Matthias Schonlau \email schonlau@uwaterloo.ca \\
       \addr Department of Statistics and Actuarial Science\\
       University of Waterloo\\
       Waterloo, Ontario, Canada}

\editor{}

\maketitle

\begin{abstract}%
Dataset distillation is a method for reducing dataset sizes by learning a small number of synthetic samples containing all the information of a large dataset. This has several benefits like speeding up model training, reducing energy consumption, and reducing required storage space. Currently, each synthetic sample is assigned a single `hard' label, and also, dataset distillation can currently only be used with image data. 

We propose to simultaneously distill both images and their labels, thus assigning each synthetic sample a `soft' label (a distribution of labels). Our algorithm increases accuracy by 2-4\% over the original algorithm for several image classification tasks. Using `soft' labels also enables distilled datasets to consist of fewer samples than there are classes as each sample can encode information for multiple classes. For example, training a LeNet model with 10 distilled images (one per class) results in over 96\% accuracy on MNIST, and almost 92\% accuracy when trained on just 5 distilled images. 

We also extend the dataset distillation algorithm to distill sequential datasets including texts. We demonstrate that text distillation outperforms other methods across multiple datasets. For example, models attain almost their original accuracy on the IMDB sentiment analysis task using just 20 distilled sentences.

Our code can be found at \url{https://github.com/ilia10000/dataset-distillation}
\end{abstract}

\begin{keywords}
  Dataset Distillation, Knowledge Distillation, Neural Networks, Synthetic Data, Gradient Descent
\end{keywords}


\begin{figure}
\centering
\includegraphics[width=0.9\textwidth]{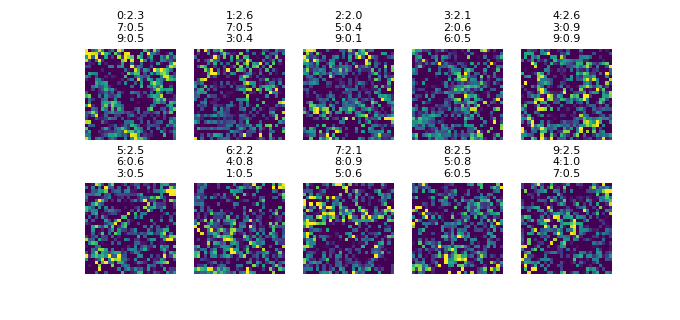}
\caption{10 MNIST images learned by SLDD can train networks with fixed initializations from $11.13\%$ distillation accuracy to $96.13\%$ ($r_{10}=97.1$). Each image is labeled with its top 3 classes and their associated logits. The full labels for these 10 images are in Table~\ref{mnist_labels}.}
\label{fig:MNIST_knowninit}
\end{figure}
\begin{table}[h!]
\caption{Learned distilled labels for the 10 distilled MNIST images in Figure~\ref{fig:MNIST_knowninit}. Distilled labels are allowed to take on any real value. If a probability distribution is needed, a softmax function can be applied to each row.}
\label{mnist_labels}
\centering
\begin{tabular}{r|rrrrrrrrrr}
                & \multicolumn{10}{c}{Digit}                                                             \\
Distilled       & 0      & 1      & 2      & 3      & 4      & 5      & 6      & 7      & 8      & 9     \\
Label           &        &        &        &        &        &        &        &        &        &       \\
\hline
1               & 2.34  & -0.33 & 0.23  & 0.04  & -0.03 & -0.23 & -0.32 & 0.54  & -0.39 & 0.49  \\
2               & -0.17 & 2.58  & 0.32  & 0.37  & -0.68 & -0.19 & -0.75 & 0.53  & 0.27  & -0.89 \\
3               & -0.26 & -0.35 & 2.00  & 0.07  & 0.08  & 0.42  & 0.02  & -0.08 & -1.09 & 0.10  \\
4               & -0.28 & 0.04  & 0.59  & 2.08  & -0.61 & -1.11 & 0.52  & 0.19  & -0.20 & 0.32  \\
5               & -0.11 & -0.52 & -0.08 & 0.90  & 2.63  & -0.44 & -0.72 & -0.39 & -0.29 & 0.87  \\
6               & 0.25  & -0.20 & -0.19 & 0.51  & -0.02 & 2.47  & 0.62  & -0.42 & -0.52 & -0.63 \\
7               & 0.42  & 0.55  & -0.09 & -1.07 & 0.83  & -0.19 & 2.16  & -0.30 & 0.26  & -0.91 \\
8               & 0.18  & -0.33 & -0.25 & 0.06  & -0.91 & 0.55  & -1.17 & 2.11  & 0.94  & 0.47  \\
9               & 0.46  & -0.48 & 0.24  & 0.09  & -0.78 & 0.75  & 0.47  & -0.40 & 2.45  & -0.71 \\
10              & -0.53 & 0.52  & -0.74 & -1.32 & 1.03  & 0.23  & 0.05  & 0.55  & 0.31  & 2.45  \\
\end{tabular}
\end{table}

\section{Introduction}
The increase in computational requirements for modern deep learning presents a range of issues. The training of deep learning models has an extremely high energy consumption~\citep{energy}, on top of the already problematic financial cost and time requirement. One path for mitigating these issues is to reduce network sizes.~\citet{kdist-hinton} proposed knowledge distillation as a method for imbuing smaller, more efficient networks with all the knowledge of their larger counterparts. Instead of decreasing network size, a second path to efficiency may be to decrease dataset size. Dataset distillation (DD) has recently been proposed as an alternative formulation of knowledge distillation to do exactly that~\citep{wang2018dataset}.
 
Dataset distillation is the process of creating a small number of synthetic samples that can quickly train a network to the same accuracy it would achieve if trained on the original dataset. It may seem counter-intuitive that training a model on a small number of synthetic images coming from a different distribution than the training data can result in comparable accuracy, but~\citet{wang2018dataset} have shown that for models with known initializations this is indeed feasible; they achieve 94\% accuracy on MNIST, a hand-written digit recognition task~\citep{mnist}, after training LeNet on just 10 synthetic images. 
\begin{figure}
\caption{\textbf{Left}: An example of a `hard' label where the second class is selected. \textbf{Center}: An example of a `soft' label restricted to being a valid probability distribution. The second class has the highest probability. \textbf{Right}: An example of an unrestricted `soft' label. The second class has the highest weight. `Hard' labels can be derived from unrestricted `soft' labels by applying the softmax function and then setting the highest probability element to 1, and the rest to 0.}
\label{soft_label_example}
\begin{align*}
  \begin{bmatrix}
           0 \\
           1 \\
           0 \\
           0 \\
           0 \\
           0 \\
           0 \\
           0 \\
           0 \\
           0 \\
\end{bmatrix}
\xleftarrow[\text{and rest to 0}]{\text{Set largest to 1}}
\begin{bmatrix}
           0.01 \\
           0.69 \\
           0.02 \\
           0.02 \\
           0.03 \\
           0.05 \\
           0.03 \\
           0.1 \\
           0.01 \\
           0.04 \\
\end{bmatrix}
\xleftarrow[\text{softmax}]{\text{Apply}}
\begin{bmatrix}
           0.8 \\
           5.1 \\
           1.5 \\
           1.5 \\
           2 \\
           2.5 \\
           2 \\
           3.2 \\
           0.8 \\
           2 \\
\end{bmatrix}
\end{align*}
\end{figure}
We propose to improve their already impressive results by learning `soft' labels as a part of the distillation process. The original dataset distillation algorithm uses fixed, or `hard', labels for the synthetic samples (e.g. the ten synthetic MNIST images each have a label corresponding to a different digit). In other words, each label is a one-hot vector: a vector where all entries are set to zero aside from a single entry, the one corresponding to the correct class, which is set to one. We relax this one-hot restriction and make the synthetic labels learnable. The resulting distilled labels are thus similar to those used for knowledge distillation as a single image can now correspond to multiple classes. An example comparing a `hard' label to a `soft' label is shown in Figure \ref{soft_label_example}. A `hard' label can be derived from a `soft' label by applying the softmax function and setting the element with the highest probability to one, while the remaining elements are set to zero.  Our soft-label dataset distillation (SLDD) not only achieves over 96\% accuracy on MNIST when using ten distilled images (as seen in Figure \ref{fig:MNIST_knowninit}), a 2\% increase over the state-of-the-art (SOTA), but also achieves almost 92\% accuracy with just five distilled images, which is less than one image per class. In addition to soft labels, we also extend dataset distillation to the natural language/sequence modeling domain and enable it to be used with several additional neural network architectures. For example, we show that Text Dataset Distillation (TDD) can train a custom convolutional neural network (CNN)~\citep{cnn} with known initialization up to 90\% of its original accuracy on the IMDB sentiment classification task~\citep{imdb} using just two synthetic sentences.  
 
The rest of this work is divided into four sections. In Section~\ref{rw}, we discuss related work in the fields of knowledge distillation, dataset reduction, and example generation. In Section~\ref{idd}, we propose improvements and extensions to dataset distillation and associated theory. In Section~\ref{exp}, we empirically validate SLDD and TDD in a wide range of experiments. Finally, in Section~\ref{con}, we discuss the significance of SLDD and TDD, and our outlook for the future. 

\section{Related Work}\label{rw}
\subsection{Knowledge Distillation}

Dataset distillation was originally inspired by network distillation~\citep{kdist-hinton} which is a form of knowledge distillation or model compression \citep{kdist-2006}. Network distillation has been studied in various contexts including when working with sequential data \citep{seq-dist}. Network distillation aims to distill the knowledge of large, or even multiple, networks into a smaller network. Similarly, dataset distillation aims to distill the knowledge of large, or even multiple, datasets into a small number of synthetic samples. `Soft' labels were recently proposed as an effective way of distilling networks by feeding the output probabilities of a larger network directly to a smaller network~\citep{kdist-hinton}, and have previously been studied in the context of different machine learning algorithms~\citep{softknn}. Our soft-label dataset distillation (SLDD) algorithm also uses `soft' labels but these are persistent and learned over the training phase of a network (rather than being produced during the inference phase as in the case of network distillation).

\subsection{Learning from `small' data}
Deep supervised learning generally requires a very large number of examples to train on. For example, MNIST and CIFAR10 both contain thousands of training images per class. Meanwhile, it appears that humans can quickly generalize from a tiny number of examples \citep{lake2015human}. Getting machines to learn from `small' data is an important aspect of trying to bridge this gap in abilities. Dataset distillation provides a method for researchers to generate synthetic examples that are optimized for allowing machines to learn from a small number of them. Studying the distilled images produced by dataset distillation may enable us to identify what allows neural networks to generalize so quickly from so few of them. In some sense, dataset distillation can be thought of as an algorithm for creating dataset summaries that machines can learn from.

\subsection{Dataset Reduction, Prototype Generation,  and Summarization}
There are a large number of methods that aim to reduce the size of a dataset with varying objectives. Active learning aims to reduce the required size of the labeled portion of a dataset by only labeling examples that are determined to be the most important~\citep{active1, active2}. Several methods aim to `prune' a dataset, or create a `core-set', by leaving in only examples that are determined to be useful~\citep{prune1, coreset1, coreset2, coreset3}. In general, all of these methods use samples from the true distribution, typically subsets of the original training set. By lifting this restriction and, instead, learning synthetic samples, dataset distillation requires far fewer samples to distill the same amount of knowledge.

In the field of nearest-neighbor classification, these dataset reduction techniques are typically referred to as `prototype selection` and `prototype generation`, and are studied extensively as methods of reducing storage requirements and improving the efficiency of nearest-neighbor classification \citep{garcia2012prototype, triguero2011taxonomy}. As with the methods above, prototype selection methods use samples from the true distribution, typically just subsets of the original training dataset. Prototype generation methods typically create samples that are not found in the training data; however, these methods are designed specifically for use with nearest-neighbor classification algorithms. 

All the dataset reduction methods discussed above also share another restriction. They all use fixed labels. Soft-label dataset distillation removes this restriction and allows the label distribution to be optimized simultaneously with the samples (or prototypes) themselves. 

\subsection{Generative Adversarial Networks}
Generative Adversarial Networks (GANs) have recently become a widely used method for image generation. They are primarily used to produce images that closely mimic those coming from the true distribution~\citep{gan1, gan2, gan3, gan4}. For dataset distillation, we instead set knowledge distillation as the objective but do not attempt to produce samples from the true distribution. Using the generator from a trained GAN may be a much faster way of producing images than the gradient-based method employed by dataset distillation. However, since the number of distilled images we aim to produce is very small, solving the objective directly through gradient-based optimization is sufficiently fast, while also more straightforward. Additionally, while some GANs can work with text~\citep{textgan1, textgan2}, they are primarily intended for image generation. 

\subsection{Measuring Problem Dimensionality}
We may intuitively believe that one deep learning task is more difficult than another. For example, when comparing the digit recognition task MNIST, to the image classification task CIFAR10~\citep{cifar}, CIFAR10 appears to be the more difficult problem though it is difficult to measure the extent of this increase in difficulty. One approach is to compare error rates for state-of-the-art (SOTA) results on these datasets. For example, the near-SOTA `dropconnect' model on MNIST achieves a 0.21\% error rate, while on CIFAR10 it achieves an error rate of 9.32\% \citep{dropconnect}. However, this approach reveals less as deeper networks approach perfect accuracy on multiple tasks.~\citet{dimension} instead derive a fairly model-independent metric for comparing the dimensionality of various problems based on the minimum number of learnable parameters needed to achieve a good local optimum. Similarly, dataset distillation aims to find the minimum number of synthetic samples needed to achieve a good local optimum. The difference is that ~\citet{dimension} constrain the number of searchable dimensions within the network weight space, while dataset distillation constrains them within the data space.

\clearpage
\section{Extending Dataset Distillation}\label{idd}
\subsection{Motivation}
As mentioned above, nearest-neighbors classification often involves data reduction techniques known as prototype selection and prototype generation. We can use these concepts, along with the k-Nearest Neighbors (kNN) classification algorithm, to visualize the difference between classical dataset reduction methods, dataset distillation, and soft-label dataset distillation. When we fit a kNN model, we essentially divide the entire space into classes based on the location of points in the training set. However, the cost of fitting a kNN model increases with the number of training points. Prototype selection and generation are methods for reducing the number of points in the training set while trying to maintain the accuracy of the original model. 

Prototype selection methods use subsets of the training set to construct the reduced training set. In the first column of Figure \ref{tab:3points}, we visualize attempts to reduce the training set for a three-class problem, the Iris flower dataset \citep{iris}, by selecting one point from each class and then fitting the kNN on these three selected points. It is clear from this visualization that only being able to use a subset of the original points, limits the ability to finely tune the resulting separation of the space. Prototype generation methods relax this restriction and create synthetic points whose placement can be optimized for resulting kNN performance. This is effectively what dataset distillation does for neural networks. In the second column of Figure \ref{tab:3points}, we generate one point for each class and then optimize the location of one of these points, before fitting the kNN on all three. Using synthetic, optimizable points allows for finer-grained tuning of the resulting class separation. In both these cases, each of the three resulting points had a single class assigned to it. 

We now propose that the three points instead be assigned an optimizable distribution of classes, effectively a `soft' label as described above. In the third column of Figure \ref{tab:3points}, in order to visualize the effect of changing a point's label distribution, we arbitrarily fix one point for each class, but we change the label distribution of one of these points, increasingly making it a mixture of the other classes, and then fit the kNN. In the final column of Figure \ref{tab:3points}, we combine the prototype generation method with our soft-label modification, to visualize the effect of simultaneously changing a point's location and label distribution. This last case is the kNN counterpart to our proposed soft-label distillation algorithm, and from the visualization, it is clear that it provides the finest-grained tuning for the resulting class separation. In fact, by using soft labels with kNN, we can separate three classes using just two points, as seen in Figure \ref{tab:2points}. 

Animated versions of both Figure \ref{tab:3points} and Figure \ref{tab:2points} are in the online appendix.
\clearpage

\begin{center}
    \begin{table}
        \centering
        \begin{tabular}{c|c|c|c}
            Selection & Generation & Soft Labels & Combined\\
             \includegraphics[scale=0.22, trim={1.7cm 1.2cm 4.7cm 1cm}, clip]{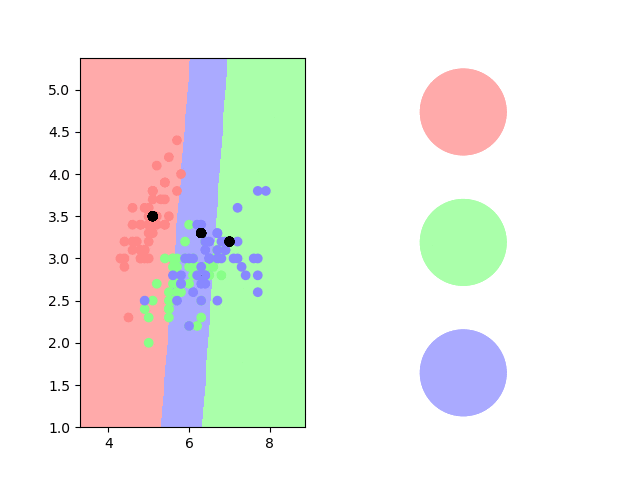} & \includegraphics[scale=0.22, trim={1.7cm 1.2cm 4.7cm 1cm}, clip]{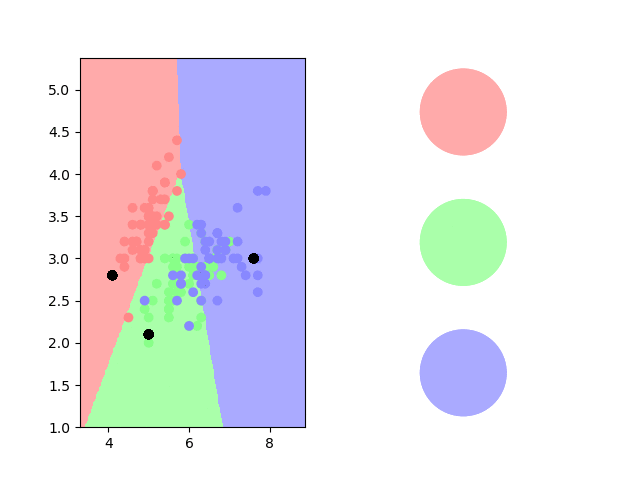} & \includegraphics[scale=0.22, trim={1.7cm 1.2cm 4.7cm 1cm}, clip]{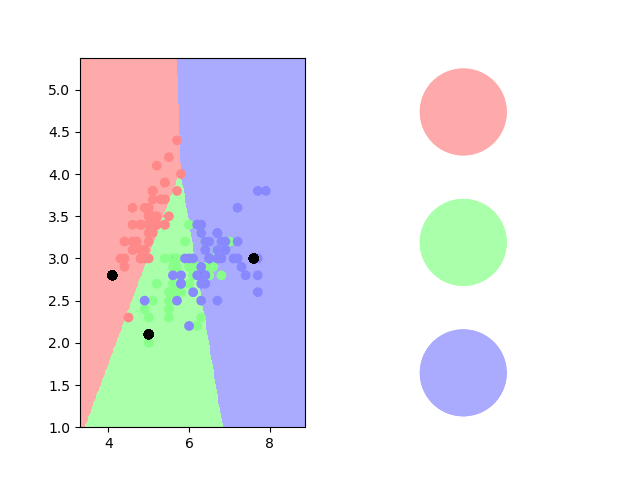} & 
             \includegraphics[scale=0.22, trim={1.7cm 1.2cm 4.7cm 1cm}, clip]{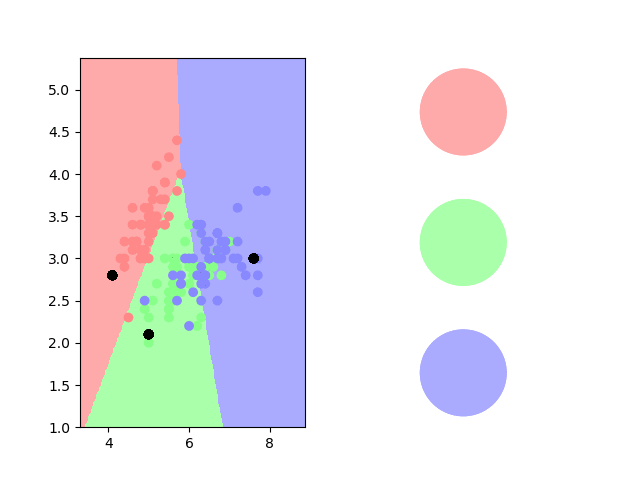}\\
             \includegraphics[scale=0.22, trim={1.7cm 1.2cm 4.7cm 1cm}, clip]{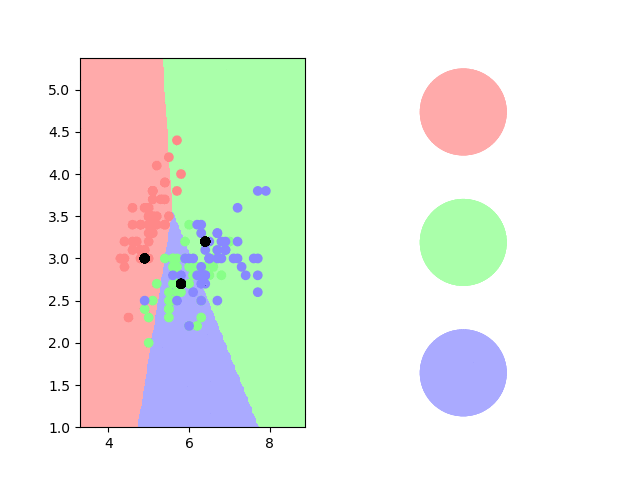} & 
             \includegraphics[scale=0.22, trim={1.7cm 1.2cm 4.7cm 1cm}, clip]{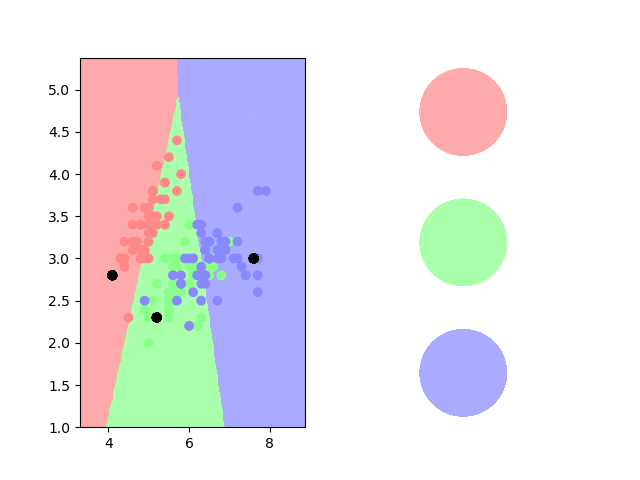} & 
             \includegraphics[scale=0.22, trim={1.7cm 1.2cm 4.7cm 1cm}, clip]{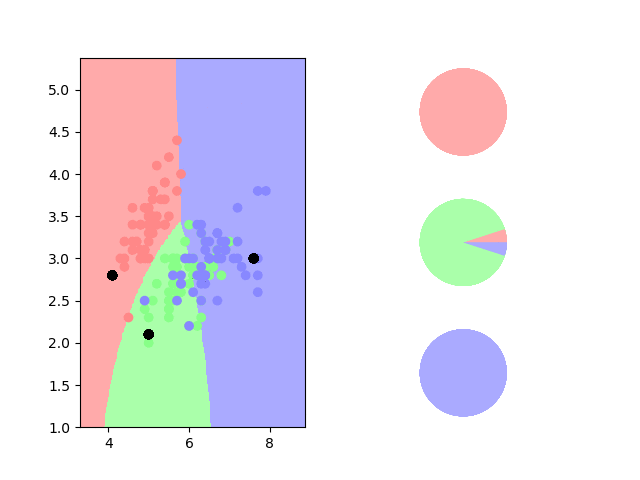} & 
             \includegraphics[scale=0.22, trim={1.7cm 1.2cm 4.7cm 1cm}, clip]{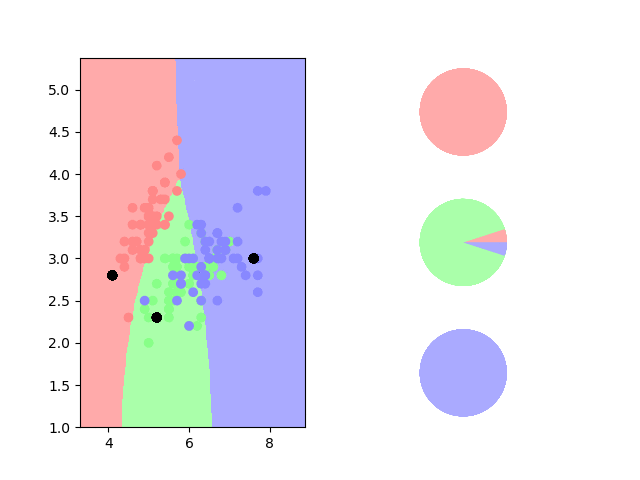} \\
             \includegraphics[scale=0.22, trim={1.7cm 1.2cm 4.7cm 1cm}, clip]{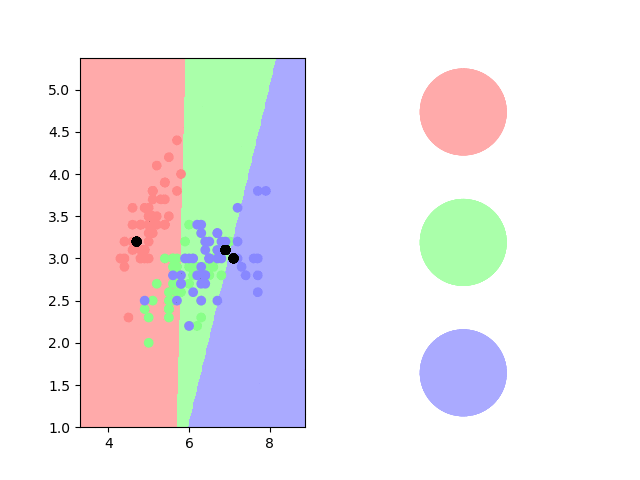} & 
             \includegraphics[scale=0.22, trim={1.7cm 1.2cm 4.7cm 1cm}, clip]{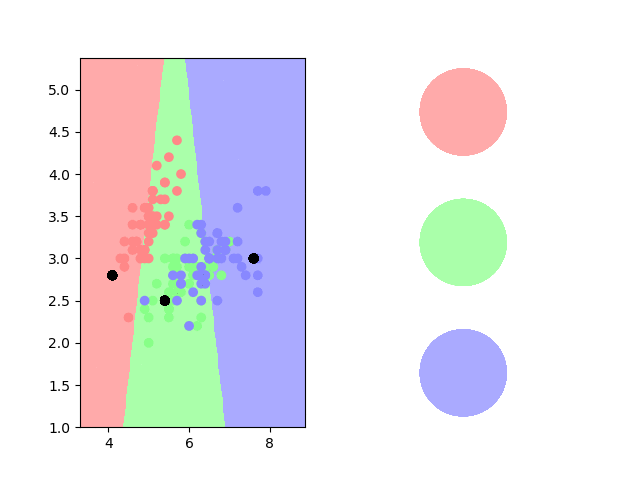} & 
             \includegraphics[scale=0.22, trim={1.7cm 1.2cm 4.7cm 1cm}, clip]{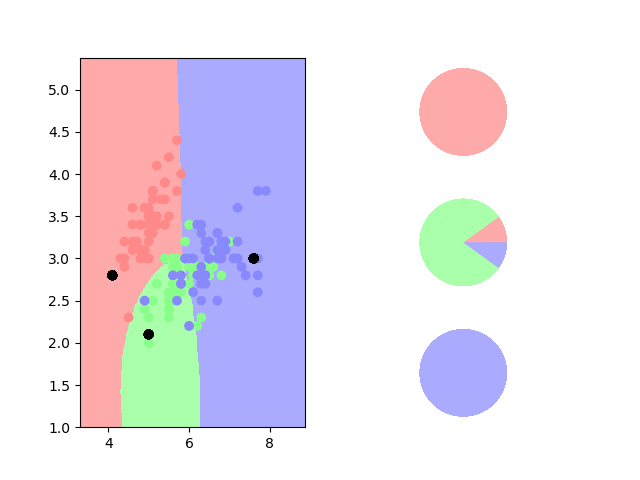} & 
             \includegraphics[scale=0.22, trim={1.7cm 1.2cm 4.7cm 1cm}, clip]{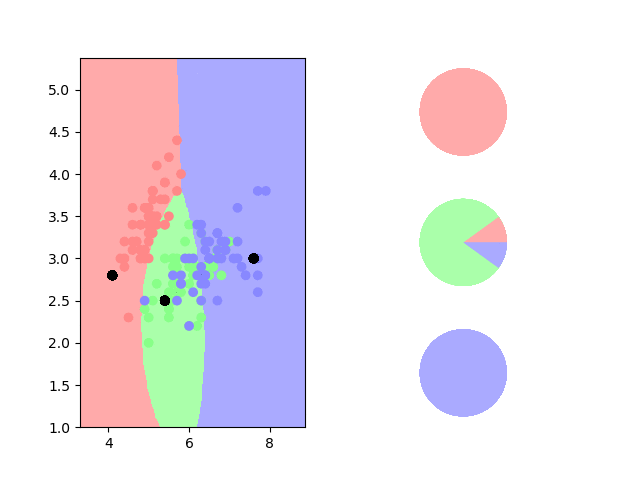}\\
             \includegraphics[scale=0.22, trim={1.7cm 1.2cm 4.7cm 1cm}, clip]{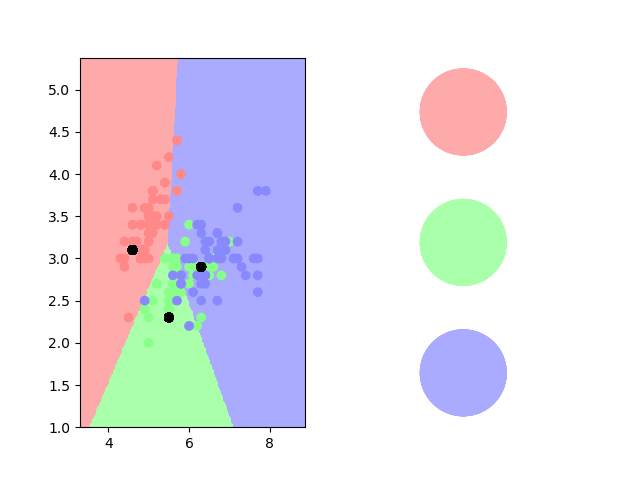} & 
             \includegraphics[scale=0.22, trim={1.7cm 1.2cm 4.7cm 1cm}, clip]{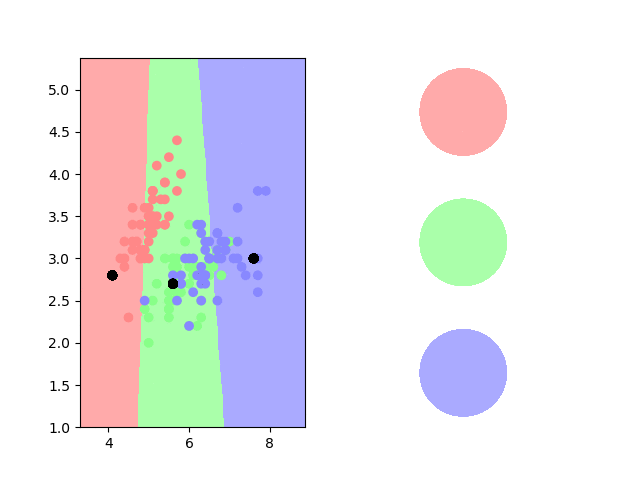} &
             \includegraphics[scale=0.22, trim={1.7cm 1.2cm 4.7cm 1cm}, clip]{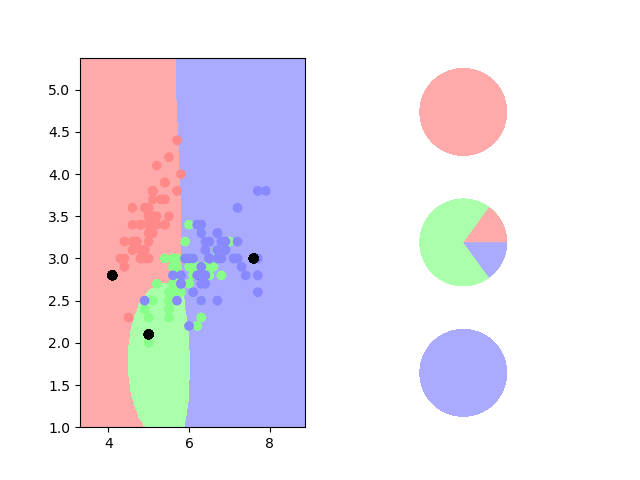} & 
             \includegraphics[scale=0.22, trim={1.7cm 1.2cm 4.7cm 1cm}, clip]{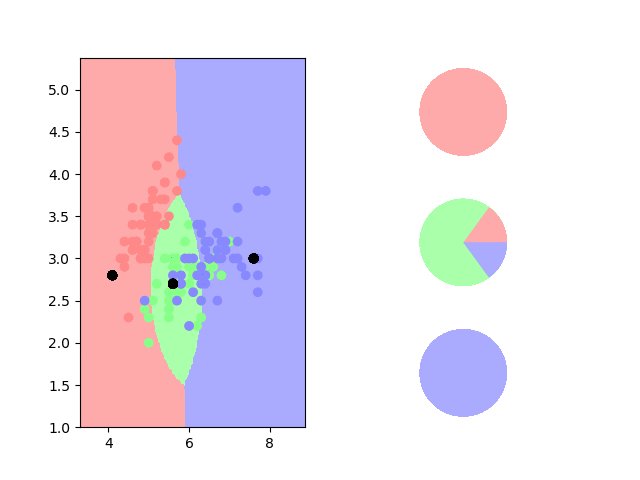}\\
        \end{tabular}
        \captionof{figure}{kNN models are fitted on 3 points obtained from the Iris flower dataset using four methods: prototype selection, prototype generation, soft labels, and prototype generation combined with soft labels. Each column contains 4 steps of the associated method used to update the 3 points used to fit the associated kNN. The pie charts represent the label distributions assigned to each of the 3 points. \textbf{Selection method:} A different random point from each class is chosen to represent its class in each of the steps. \textbf{Generation method:} The middle point associated with the 'green' label is moved diagonally in each step. \textbf{Soft labels method:} The label distribution of the middle point is changed each step to contain a larger proportion of both other classes. \textbf{Combined method:} The middle point is simultaneously moved and has its label distribution updated in each step. }
        \label{tab:3points}
    \end{table}
\end{center}
\clearpage
\begin{center}
    \begin{table}
        \centering
        \begin{tabular}{cccc}
             \includegraphics[scale=0.32, trim={1.1cm 0.5cm 3.2cm 0.5cm}, clip]{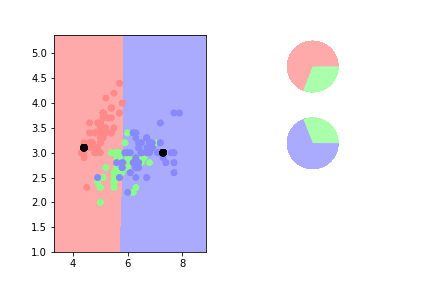}
             & \includegraphics[scale=0.32, trim={1.1cm 0.5cm 3.2cm 0.5cm}, clip]{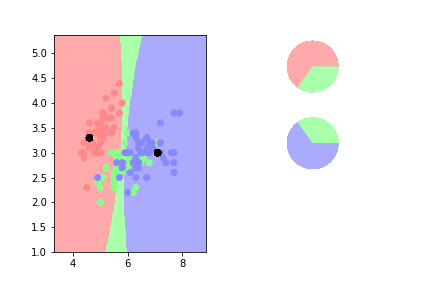} 
             & \includegraphics[scale=0.32, trim={1.1cm 0.5cm 3.2cm 0.5cm}, clip]{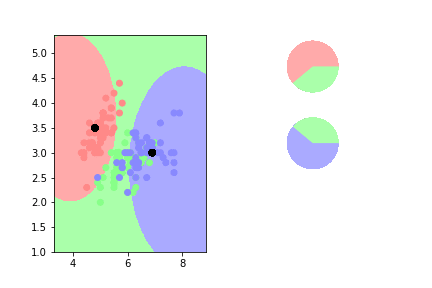}
             & \includegraphics[scale=0.32, trim={1.1cm 0.5cm 3.2cm 0.5cm}, clip]{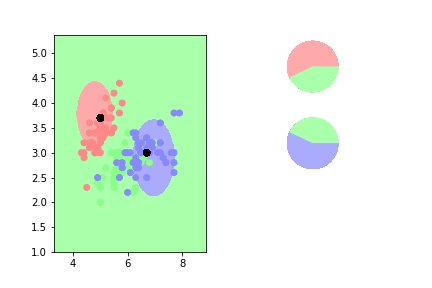} \\
        \end{tabular}
        \captionof{figure}{kNN model fitted on 2 points obtained using a combination of prototype generation and soft labels. The pie charts represent the label distributions assigned to each of the 2 points. From left-to-right, in each plot, the locations of the 2 points are slightly shifted and the values associated with their `green` label are increased. By modifying the location and soft labels of the 2 points, the space can still be separated into 3 classes.}
        \label{tab:2points}
    \end{table}
\end{center}
\subsection{Basic Approach}
Our basic approach is the same as \citet{wang2018dataset}. We summarize it here in a slightly modified way to explicitly show the labels of the distilled dataset. This additional notation becomes useful once we enable label learning in the next section. 

Given a training dataset $\mathbf{d}=\left\{x_{i}, y_{i}\right\}_{i=1}^{N}$, a neural network with parameters $\theta$, and a twice-differentiable loss function $\ell\left(x_{i}, y_{i}, \theta\right)$, our objective is to find
\begin{equation}
    \theta^{*}=\underset{\theta}{\arg \min } \frac{1}{N} \sum_{i=1}^{N} \ell\left(x_{i},y_{i}, \theta\right) \triangleq \underset{\theta}{\arg \min }\: \ell(\mathbf{x}, \mathbf{y}, \theta)\text{ .}
\end{equation}

In general, training with stochastic gradient descent (SGD) involves repeatedly sampling mini-batches of training data and updating network parameters by their error gradient scaled by learning rate $\eta$.
\begin{equation}
    \theta_{t+1}=\theta_{t}-\eta \nabla_{\theta_{t}} \ell\left(\mathbf{x}_{t}, \mathbf{y}_{t}, \theta_{t}\right)
\end{equation}

With dataset distillation, the goal is to perform just one such step while still achieving the same accuracy. We do this by learning a very small number of synthetic samples $\tilde{\mathbf{x}}$ that minimize $\mathcal{L}$, a one-step loss objective, for $\theta_{1}=\theta_{0}-\tilde{\eta} \nabla_{\theta_{0}} \ell\left(\tilde{\mathbf{x}}, \theta_{0}\right)$.
\begin{equation}
    \mathcal{L}\left(\tilde{\mathbf{x}}, \tilde{\mathbf{y}}, \tilde{\eta} ; \theta_{0}\right):=\ell\left(\mathbf{x}, {\mathbf{y}}, \theta_{1}\right)=\ell\left(\mathbf{x}, {\mathbf{y}}, \theta_{0}-\tilde{\eta} \nabla_{\theta_{0}} \ell\left(\tilde{\mathbf{x}}, \tilde{\mathbf{y}}, \theta_{0}\right)\right)
\end{equation}
\begin{equation}
    \label{L}
    \tilde{\mathbf{x}}^{*}, \tilde{\eta}^{*}=\underset{\tilde{\mathbf{x}}, \tilde{\eta}}{\arg \min{\mathcal{L}}}\left(\tilde{\mathbf{x}}, \tilde{\mathbf{y}}, \tilde{\eta} ; \theta_{0}\right)=\underset{\tilde{\mathbf{x}}, \tilde{\eta}}{\arg \min{\ell}}\left(\mathbf{x}, {\mathbf{y}}, \theta_{0}-\tilde{\eta} \nabla_{\theta_{0}} \ell\left(\tilde{\mathbf{x}}, \tilde{\mathbf{y}}, \theta_{0}\right)\right)
\end{equation} 
Note that, currently, we are minimizing over $\tilde{\mathbf{x}} \text{ and } \tilde{\eta}$, but not $\tilde{\mathbf{y}}$, as the distilled labels are fixed for the original dataset distillation algorithm. We minimize this objective, or in other words `learn the distilled samples', by using standard gradient descent.

\subsection{Learnable Labels}
As mentioned above, one formulation of knowledge distillation proposes that a smaller network be trained on the outputs of a larger network rather than the original training labels. Unlike the training labels, the output labels of the larger network are not `hard' labels. Because they are generally the outputs of a softmax layer, the output labels form a probability distribution over the possible classes. The idea is that any training image contains information about more than one class (e.g. an image of the digit `3' looks a lot like other digits `3' but it also looks like the digit `8'). Using `soft' labels allows us to convey more information about the associated image.

The original dataset distillation algorithm was restricted to `hard' labels for the distilled data; each distilled image has to be associated with just a single class. We relax this restriction and allow distilled labels to take on any real value. Since the distilled labels are now continuous variables, we can modify the distillation algorithm in order to make the distilled labels learnable using the same method as for the distilled images: a combination of backpropagation and gradient descent. With our modified notation, we simply need to change equation (\ref{L}) to also minimize over $\tilde{\mathbf{y}}$.
\begin{equation}
    \tilde{\mathbf{x}}^{*}, \tilde{\mathbf{y}}^{*}, \tilde{\eta}^{*}=\underset{\tilde{\mathbf{x}}, \tilde{\mathbf{y}}, \tilde{\eta}}{\arg \min{\mathcal{L}}}\left(\tilde{\mathbf{x}}, \tilde{\mathbf{y}}, \tilde{\eta} ; \theta_{0}\right)=\underset{\tilde{\mathbf{x}}, \tilde{\eta}}{\arg \min{\ell}}\left(\mathbf{x}, {\mathbf{y}}, \theta_{0}-\tilde{\eta} \nabla_{\theta_{0}} \ell\left(\tilde{\mathbf{x}}, \tilde{\mathbf{y}}, \theta_{0}\right)\right)
\end{equation} 
Algorithm~\ref{SLDDalgo} details this soft-label dataset distillation (SLDD) algorithm. We note that in our experiments, we generally initialize $\tilde{y}$ with the one-hot values that `hard' labels would have. We found that this tends to increase accuracy when compared to random initialization, perhaps because it encourages more differentiation between classes early on in the distillation process.

\begin{algorithm}
  \caption{Soft-Label Dataset Distillation (SLDD)}
  \label{SLDDalgo}
  \textbf{Input:} $p(\theta_0)$: distribution of initial weights; $M$: the number of distilled data; $\alpha$: step size; $n$: batch size; $T$: number of optimization iterations; $\tilde{y}_0$: initial value for $\tilde{y}$; $\tilde{\eta}_0$: initial value for $\tilde{\eta}$\\
  \begin{algorithmic}[1]
    \STATE{Initialize distilled data\\ $\tilde{\mathbf{x}}=\left\{\tilde{x}_{i}\right\}_{i=1}^{M}$ randomly, \\ $\tilde{\mathbf{y}}=\left\{\tilde{y}_{i}\right\}_{i=1}^{M}$ $\leftarrow \tilde{y}_{0}$, \\$\tilde{\eta} \leftarrow \tilde{\eta}_{0}$ }
    \FOR{each training step t = 1 to T}
        \STATE Get a mini-batch of real training data\\ $(\mathbf{x}_{t},\mathbf{y}_{t})=\left\{x_{t, j},y_{t, j}\right\}_{j=1}^{n}$
        \STATE One-hot encode the labels\\ $(\mathbf{x}_{t},\mathbf{y^*}_{t})=\left\{x_{t, j},\text{Encode}(y_{t, j})\right\}_{j=1}^{n}$
        
        \STATE Sample a batch of initial weights \\ $\theta_{0}^{(j)} \sim p\left(\theta_{0}\right)$
        \FOR{each sampled $\theta_{0}^{(j)}$}
            \STATE Compute updated model parameter with GD\\ $\theta_{1}^{(j)}=\theta_{0}^{(j)}-\tilde{\eta} \nabla_{\theta_{0}^{(j)}} \ell\left(\tilde{\mathbf{x}},\tilde{\mathbf{y}}, \theta_{0}^{(j)}\right)$
            \STATE Evaluate the objective function on real training data: $\mathcal{L}^{(j)}=\ell\left(\mathbf{x}_{t},\mathbf{y^*}_{t}, \theta_{1}^{(j)}\right)$
        \ENDFOR
        \STATE Update distilled data\\ $\tilde{\mathbf{x}} \leftarrow \tilde{\mathbf{x}}-\alpha \nabla_{\tilde{\mathbf{x}}} \sum_{j} \mathcal{L}^{(j)},$\\ $\tilde{\mathbf{y}} \leftarrow \tilde{\mathbf{y}}-\alpha \nabla_{\tilde{\mathbf{y}}} \sum_{j} \mathcal{L}^{(j)},$ and \\ $\tilde{\eta} \leftarrow \tilde{\eta}-\alpha \nabla_{\tilde{\eta}} \sum_{j} \mathcal{L}^{(j)}$ 
    \ENDFOR
  \end{algorithmic}
  \textbf{Output:} distilled data $\tilde{\mathbf{x}}$; distilled labels $\tilde{\mathbf{y}}$; optimized learning rate $\tilde{\eta}$
\end{algorithm}

\subsection{Text and Other Sequences}
The original dataset distillation algorithm was only shown to work with image data, but intuitively, there is no reason why text or other sequences should not be similarly distillable. However, it is difficult to use gradient methods directly on text data as it is discrete. In order to be able to use SLDD with text data, we need to first embed the text data into a continuous space. This is a common practice when working with many modern natural language processing models, though the embedding method itself can vary greatly \citep{ma2016end,devlin2018bert,peters2018deep}. Any popular embedding method can be used; in our experiments, we used pre-trained GloVe embeddings~\citep{glove}. Once the text is embedded into a continuous space, the problem of distilling it becomes analogous to soft-label image distillation. If all sentences are padded/truncated to some pre-determined length, then each sentence is essentially just a one-channel image of size [length]$*$[embedding dimension]. When working with models that do not require fixed-length input, like recurrent neural networks (RNN), the distilled sentences do not have to be padded/truncated to the same length. It is also important to note that the embedding is performed only on sentences coming from the true dataset; the distilled samples are learned directly as embedded representations. Since the distilled data produced by this algorithm would still be in the embedding space, it may be of interest to find the nearest sentences that correspond to the distilled embeddings. To compute the nearest sentence to a distilled embedding matrix, for every column vector in the matrix, the nearest embedding vector from the original dictionary must be found. These embedding vectors must then be converted back into their corresponding words, and those words joined into a sentence.
The resulting algorithm for text dataset distillation (TDD) is detailed in Algorithm~\ref{TDDalgo} which is a modification of the SLDD Algorithm~\ref{SLDDalgo}.

\begin{algorithm}
  \caption{Text Dataset Distillation (TDD)}
  \label{TDDalgo}
  \textbf{Input:} $p(\theta_0)$: distribution of initial weights; $M$: the number of distilled data; $\alpha$: step size; $n$: batch size; $T$: number of optimization iterations; $\tilde{y}_0$: initial value for $\tilde{y}$; $\tilde{\eta}_0$: initial value for $\tilde{\eta}$; $s$: sentence length; $d$: embedding size\\
  \begin{algorithmic}[1]
    \STATE{Initialize distilled data\\ $\tilde{\mathbf{x}}=\left\{\tilde{x}_{i}\right\}_{i=1}^{M}$ randomly of size $s{\times}d$, \\ $\tilde{\mathbf{y}}=\left\{\tilde{y}_{i}\right\}_{i=1}^{M}$ $\leftarrow \tilde{y}_{0}$, \\$\tilde{\eta} \leftarrow \tilde{\eta}_{0}$ }
    \FOR{each training step t = 1 to T}
        \STATE Get a mini-batch of real training data\\ $(\mathbf{x}_{t},\mathbf{y}_{t})=\left\{x_{t, j},y_{t, j}\right\}_{j=1}^{n}$
        \STATE Pad (or truncate) each sentence in the mini-batch\\
        $(\mathbf{x^p}_{t},\mathbf{y}_{t})=\left\{\text{Pad}(x_{t, j}, \text{len} = s),y_{t, j}\right\}_{j=1}^{n}$
        \STATE Embed each sentence in the mini-batch\\
        $(\mathbf{x^*}_{t},\mathbf{y}_{t})=\left\{\text{Embed}(x^p_{t, j}, \text{dim} = d),y_{t, j}\right\}_{j=1}^{n}$
        \STATE One-hot encode the labels\\ $(\mathbf{x^*}_{t},\mathbf{y^*}_{t})=\left\{x^*_{t, j},\text{Encode}(y_{t, j})\right\}_{j=1}^{n}$
        \STATE Sample a batch of initial weights \\ $\theta_{0}^{(j)} \sim p\left(\theta_{0}\right)$
        \FOR{each sampled $\theta_{0}^{(j)}$}
            \STATE Compute updated model parameter with GD\\ $\theta_{1}^{(j)}=\theta_{0}^{(j)}-\tilde{\eta} \nabla_{\theta_{0}^{(j)}} \ell\left(\tilde{\mathbf{x}},\tilde{\mathbf{y}}, \theta_{0}^{(j)}\right)$
            \STATE Evaluate the objective function on real training data: $\mathcal{L}^{(j)}=\ell\left(\mathbf{x^*}_{t},\mathbf{y^*}_{t}, \theta_{1}^{(j)}\right)$
        \ENDFOR
        \STATE Update distilled data\\ $\tilde{\mathbf{x}} \leftarrow \tilde{\mathbf{x}}-\alpha \nabla_{\tilde{\mathbf{x}}} \sum_{j} \mathcal{L}^{(j)},$\\ $\tilde{\mathbf{y}} \leftarrow \tilde{\mathbf{y}}-\alpha \nabla_{\tilde{\mathbf{y}}} \sum_{j} \mathcal{L}^{(j)},$ and \\ $\tilde{\eta} \leftarrow \tilde{\eta}-\alpha \nabla_{\tilde{\eta}} \sum_{j} \mathcal{L}^{(j)}$ 
    \ENDFOR
    \FOR {$i = 1 \text{ to } M$}
    \STATE Compute nearest embedding for every distilled word \\ 
    $\tilde{\mathbf{x}}^*_i = \left\{\text{NearestEmbed}(\tilde{x}_{i, j})\right\}_{j=1}^{s}$
    \STATE Decode embedding into text\\
    $\tilde{\mathbf{z}}_i = \left\{\text{WordFromEmbed}(\tilde{x}^*_{i, j})\right\}_{j=1}^{s}$
    \ENDFOR
    \STATE $\tilde{\mathbf{z}}=\left\{\tilde{z}_{i}\right\}_{i=1}^{M}$
  \end{algorithmic}
  \textbf{Output:} distilled data $\tilde{\mathbf{x}}$; distilled labels $\tilde{\mathbf{y}}$; optimized learning rate $\tilde{\eta}$; nearest sentences $\tilde{\mathbf{z}}$
\end{algorithm}

\subsection{Random initializations and multiple steps}
\begin{figure}
\centering
\begin{subfigure}{1\textwidth}
\subcaption{Step 0}
\centering
\includegraphics[width=0.8\textwidth]{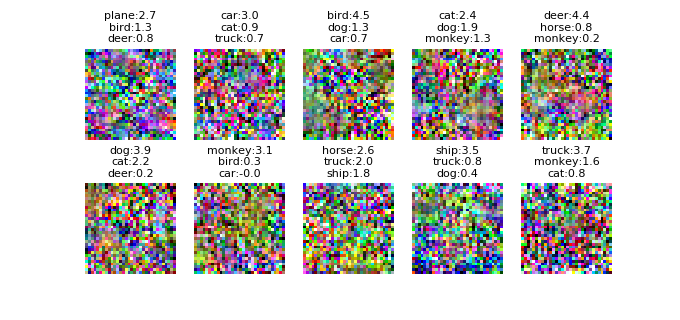}
\end{subfigure}
\begin{subfigure}{1\textwidth}
\subcaption{Step 5}
\centering
\includegraphics[width=0.8\textwidth]{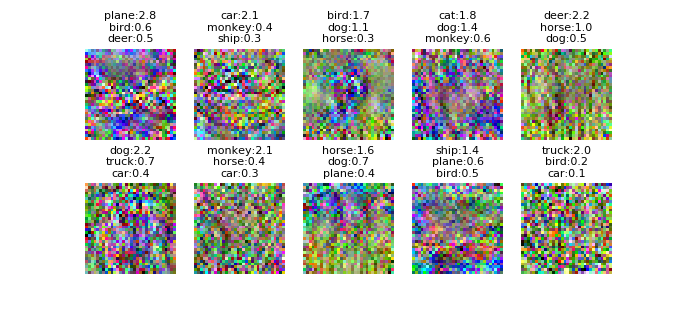}
\end{subfigure}
\begin{subfigure}{1\textwidth}
\subcaption{Step 9}
\centering
\includegraphics[width=0.8\textwidth]{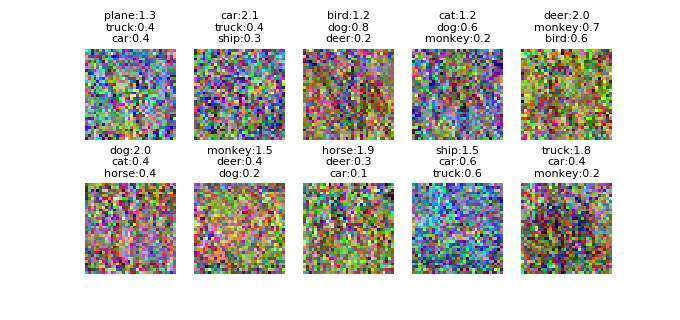}
\end{subfigure}
\caption{SLDD can learn 100 distilled CIFAR10 images that train networks with fixed initializations from $12.9\%$ distillation accuracy to $60.0\%$ ($r_{100}=75.0$). Each image is labeled with its top 3 classes and their associated logits. Only 3 of the 10 steps are shown.}
\label{fig:CIFAR_knowninit}
\end{figure}

The procedures we described above make one important assumption: network initialization $\theta_0$ is fixed. The samples created this way do not lead to high accuracies when the network is re-trained on them with a different initialization as they contain information not only about the dataset but also about $\theta_0$. In the distilled images in Figures~\ref{fig:MNIST_knowninit} and ~\ref{fig:CIFAR_knowninit}, this can be seen as what looks like a lot of random noise. \citet{wang2018dataset} propose that the method instead be generalized to work with network initializations randomly sampled from some restricted distribution. 
\begin{equation}
    \tilde{\mathbf{x}}^{*}, \tilde{\mathbf{y}}^{*}, \tilde{\eta}^{*}=\underset{\tilde{\mathbf{x}},\tilde{\mathbf{y}}, \tilde{\eta}}{\arg \min }\: \mathbb{E}_{\theta_{0} \sim p\left(\theta_{0}\right)} \mathcal{L}\left(\tilde{\mathbf{x}},\tilde{\mathbf{y}}, \tilde{\eta} ; \theta_{0}\right)
\end{equation} The resulting images, especially for MNIST, appear to have much clearer patterns and much less random noise, and the results detailed in Section~\ref{exp} suggest that this method generalizes fairly well to other randomly sampled initializations from the same distribution.

Additionally, \citet{wang2018dataset} suggest that the above methods can work with multiple gradient descent (GD) steps. If we want to perform multiple gradient descent steps, each with a different mini-batch of distilled data, we simply backpropagate the gradient through every one of these additional steps. Finally, it may also be beneficial to train the neural networks on the distilled data for more than one epoch. The experimental results suggest that multiple steps and multiple epochs improve distillation performance for both image and text data, particularly when using random network initializations. 

\section{Experiments}\label{exp}
\subsection{Metrics}
The simplest metric for gauging distillation performance is to train a model on distilled samples and then test it on real samples. We refer to the accuracy achieved on these real samples as the `distillation accuracy'. 
However, several of the models we use in our experiments do not achieve SOTA accuracy on the datasets they are paired with, so it is useful to construct a relative metric that compares distillation accuracy to original accuracy. The first such metric is the `distillation ratio' which we define as the ratio of distillation accuracy to original accuracy. The distillation ratio is heavily dependent on the number of distilled samples so the notation we use is $r_M = 100\%*\frac{[\text{distillation accuracy}]}{[\text{original accuracy}]}, M=[\text{number of distilled samples}]$. We may refer to this metric as the `$M$-sample distillation ratio' when clarification is needed. It may also be of interest to find the minimum number of distilled images required to achieve a certain distillation ratio. To this end we can define a second relative metric that we call the `$A$\% distillation size', and we write $d_A=M$ where $M$ is the minimum number of distilled samples required to achieve a distillation ratio of $A$\%. 

\subsection{Image Data}
The LeNet model we use with MNIST achieves nearly SOTA results, 99\% accuracy, so it is sufficient to use distillation accuracy when describing distillation performance with it. However, AlexCifarNet only achieves 80\% on CIFAR10 so it is helpful to use the 2 relative metrics when describing this set of distillation results.

\paragraph{Baselines.}
It is useful to compare dataset distillation against several other methods of dataset reduction. We use the following baselines suggested by~\citet{wang2018dataset}.
\begin{itemize}
    \item \textbf{Random real images:} We randomly sample the same number of real images per class from the training data. These images are used for two baselines: training neural networks and training K-Nearest Neighbors classifiers.
    \item \textbf{Optimized real images:} We sample several sets of random real images as above, but now we choose the 20\% of these sets that have the best performance on training data. These images are used for one baseline: training neural networks.
    \item \textbf{$k$-means:} We use $k$-means to learn clusters for each class, and keep the resulting centroids. These images are used for two baselines: training neural networks and training K-Nearest Neighbors classifiers.
    \item \textbf{Average real images:} We compute the average image for each class and use it for training. These images are used for one baseline: training neural networks.
\end{itemize}
Each of these baseline methods produces a small set of images that can be used to train models. All four of the baseline methods are used to train and test LeNet and AlexCifarNet on their respective datasets. Additionally, two of the baseline methods are used to also train K-Nearest Neighbor classifiers to compare performance against neural networks. The results for these six baselines, as determined by~\citet{wang2018dataset}, are shown in Table~\ref{tab:img_results}.

\begin{figure}
  \includegraphics[width=1\textwidth]{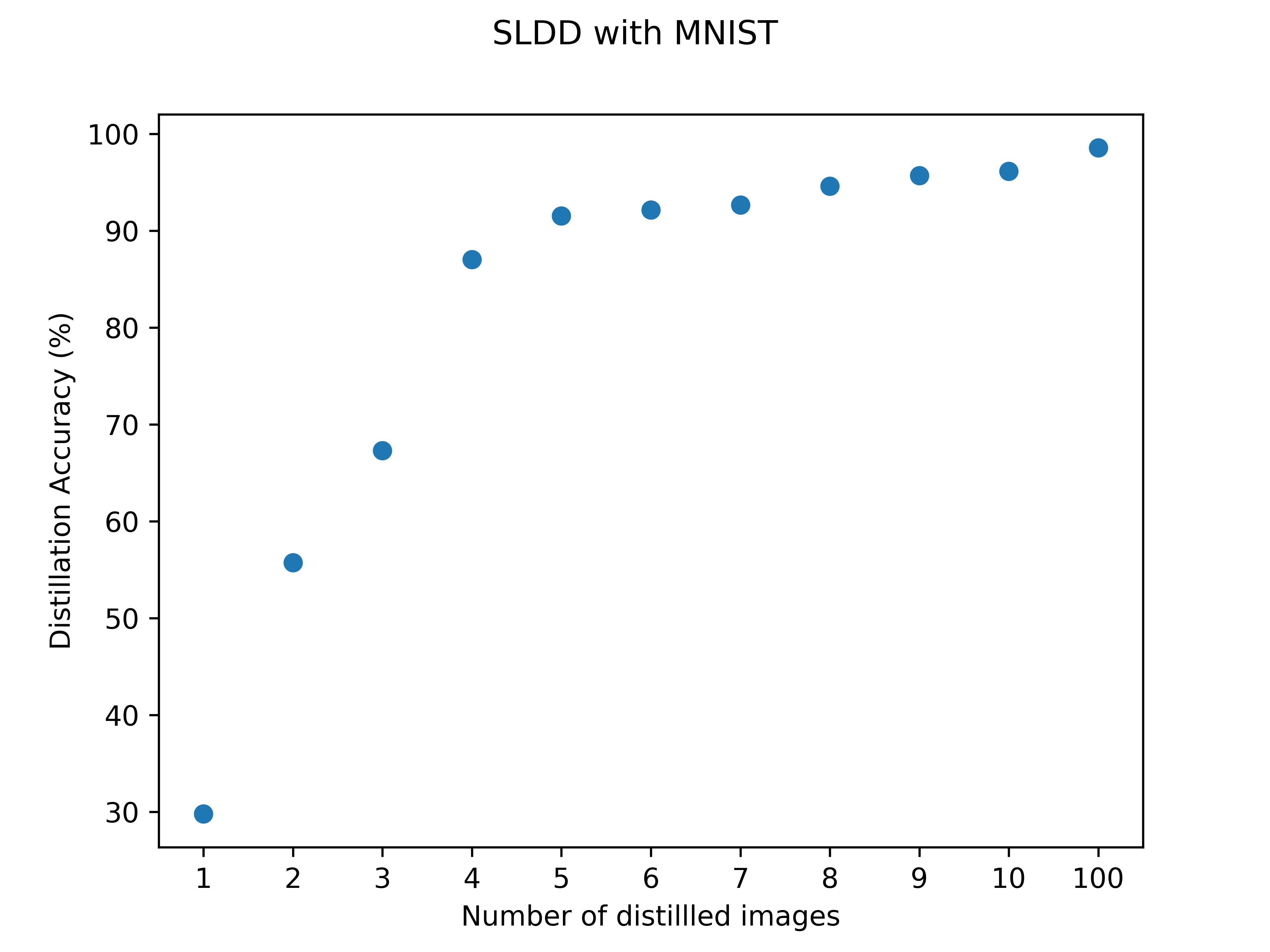}
\caption{Distillation accuracy on MNIST with LeNet for different distilled dataset sizes.}
\label{fig:MNIST_compress}       
\end{figure}

\paragraph{Fixed initialization.}
When the network initialization is fixed between the distillation and training phases, synthetic images produced by dataset distillation result in high distillation accuracies. The SLDD algorithm produces images that result in equal or higher accuracies when compared to the original DD algorithm. For example, DD can produce 10 distilled images that train a LeNet model up to 93.76\% accuracy on MNIST~\citep{wang2018dataset}. Meanwhile, SLDD can produce 10 distilled images that train the same model up to 96.13\% accuracy (Figure~\ref{fig:MNIST_knowninit}). The full distilled labels for these 10 images are laid out in Table~\ref{mnist_labels}. SLDD can even produce a tiny set of just 5 distilled images that train LeNet to 91.56\% accuracy.  As can be seen in Figure~\ref{fig:MNIST_compress}, the 90\% distillation size (i.e. the minimum number of images needed to achieve 90\% of the original accuracy) of MNIST with fixed initializations is $d_A=5$, and while adding more distilled images typically increases distillation accuracy, this begins to plateau after five images. Similarly, SLDD provides a 7.5\% increase in 100-sample distillation ratio (6\% increase in distillation accuracy) on CIFAR10 over DD. Based on these results, detailed further in Table~\ref{tab:img_results}, it appears that SLDD is even more effective than DD at distilling image data into a small number of samples. This intuitively makes sense as the learnable labels used by SLDD increase the capacity of the distilled dataset for storing information. 

\begin{figure}
\centering
\begin{subfigure}{1\textwidth}
\subcaption{Step 0}
\centering
\includegraphics[width=0.8\textwidth]{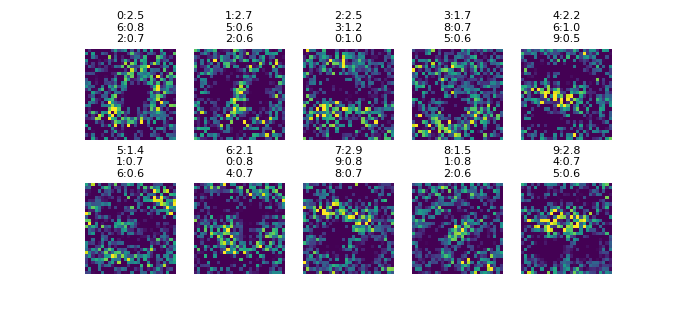}
\end{subfigure}
\begin{subfigure}{1\textwidth}
\subcaption{Step 5}
\centering
\includegraphics[width=0.8\textwidth]{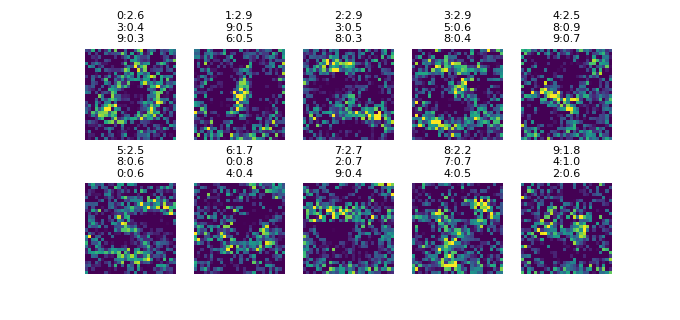}
\end{subfigure}
\begin{subfigure}{1\textwidth}
\subcaption{Step 9}
\centering
\includegraphics[width=0.8\textwidth]{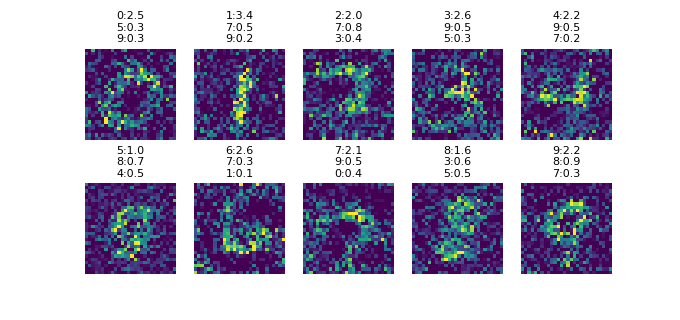}
\end{subfigure}
\caption{SLDD can learn 100 distilled MNIST images that train networks with random initializations from $10.09\% \pm 2.54\%$ distillation accuracy to $82.75\% \pm 2.75\%$ ($r_{100}=83.6$). Each image is labeled with its top 3 classes and their associated logits. Only 3 of the 10 steps are shown.}
\label{fig:MNIST_unkinit}
\end{figure}

\begin{figure}
\centering
\begin{subfigure}{1\textwidth}
\centering
\subcaption{Step 0}
\includegraphics[width=0.8\textwidth]{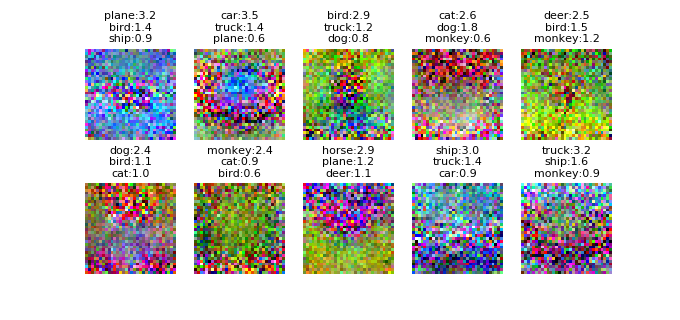}
\end{subfigure}
\begin{subfigure}{1\textwidth}
\centering
\subcaption{Step 5}
\includegraphics[width=0.8\textwidth]{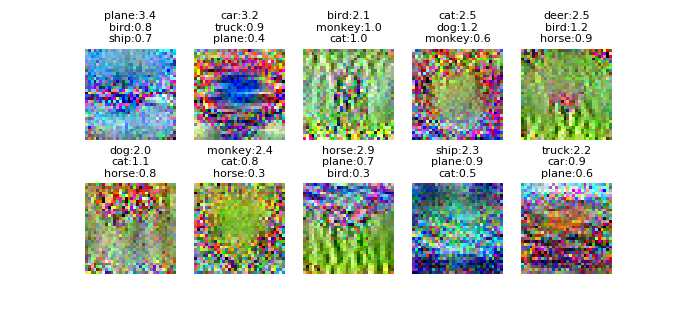}
\end{subfigure}
\begin{subfigure}{1\textwidth}
\centering
\subcaption{Step 9}
\includegraphics[width=0.8\textwidth]{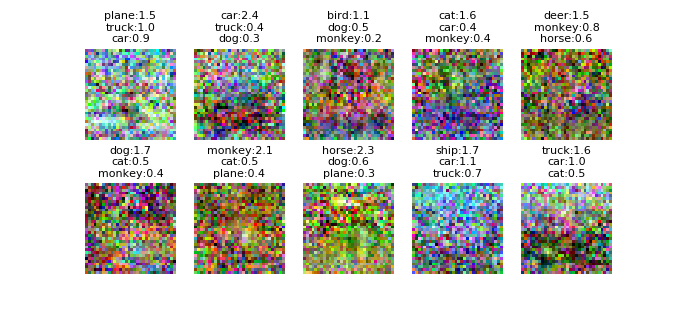}
\end{subfigure}
\caption{SLDD can learn 100 distilled CIFAR10 images that train networks with random initializations from $10.17\% \pm 1.23\%$ distillation accuracy to $39.82\% \pm 0.83\%$ ($r_{100}=49.8$). Each image is labeled with its top 3 classes and their associated logits. Only 3 of the 10 steps are shown.}
\label{fig:CIFAR_unkinit}
\end{figure}

\paragraph{Random initialization.}
It is also of interest to know whether distilled data are robust to network initialization. Specifically, we aim to identify if distilled samples store information only about the network initializations, or whether they can store information contained within the training data. To this end, we perform experiments by sampling random network initializations generated using the Xavier Initialization~\citep{xavier}. The distilled images produced in this way are more representative of the training data but generally result in lower accuracies when models are trained on them. Once again, images distilled using SLDD lead to higher distillation accuracies than DD when the number of distilled images is held constant. For example, 100 MNIST images learned by DD result in accuracies of 79.5 $\pm$ 8.1\%, while 100 images learned by SLDD result in accuracies of 82.75 $\pm$ 2.75\%. There is similarly a 3.8\% increase in 100-sample distillation ratio (3\% increase in distillation accuracy) when using SLDD instead of DD on CIFAR10 using 100 distilled images each. These results are detailed in Table~\ref{tab:img_results}. It is also interesting to note that the actual distilled images, as seen in Figures~\ref{fig:MNIST_unkinit} and~\ref{fig:CIFAR_unkinit}, appear to have much clearer patterns emerging than in the fixed initialization case. These results suggest that DD, and even more so SLDD, can be generalized to work with random initializations and distill knowledge about the dataset itself when they are trained this way. All the mean and standard deviation results for random initializations in Table~\ref{tab:img_results} are derived by testing with 200 randomly initialized networks. 

\begin{sidewaystable}
\caption{Means and standard deviations of distillation and baseline accuracies on image data. All values are percentages. The first four baselines are used to train the same neural network as in the distillation experiments. The last two baselines are used to train a K-Nearest Neighbors classifier. Experiments with random initializations have their results listed in the form [mean $\pm$ standard deviation] and are based on the resulting performance of 200 randomly initialized networks.}
\label{tab:img_results}
\centering
\begin{tabular}{lllllllllll}
        \hline
        & \multicolumn{2}{c}{SLDD accuracy} & \multicolumn{2}{c}{DD accuracy} & \multicolumn{4}{c}{Used as training data in same \# of GD steps} &  \multicolumn{2}{c}{Used in K-NN} \\
        & Fixed & Random & Fixed & Random &  Rand. real & Optim. real & $k$-means & Avg. real  &  Rand. real & k-means\\
        \hline
        MNIST & \textbf{98.6} & 82.7 $\pm$ 2.8 & 96.6 & 79.5 $\pm$ 8.1 & 68.6 $\pm$ 9.8 & 73.0 $\pm$ 7.6 & 76.4 $\pm$ 9.5 & 77.1 $\pm$ 2.7 & 71.5 $\pm$ 2.1 & \textbf{92.2 $\pm$ 0.1}\\
        CIFAR10 & \textbf{60.0} & \textbf{39.8 $\pm$ 0.8} & 54.0 & 36.8 $\pm$ 1.2 & 21.3 $\pm$ 1.5 & 23.4 $\pm$ 1.3 & 22.5 $\pm$ 3.1 & 22.3 $\pm$ 0.3 & 18.8 $\pm$ 1.3 & 29.4 $\pm$ 0.3\\
        \hline
\end{tabular}

\vspace{3\baselineskip}

\caption{Means and standard deviations of TDD and baseline accuracies on text data using TextConvNet. All values are percentages. The first four baselines are used to train the same neural network as in the distillation experiments. The last two baselines are used to train a K-Nearest Neighbors classifier. Each result uses 10 GD steps aside from IMDB with $k$-means and TREC50  which had to be done with 2 GD steps due to GPU memory constraints and also insufficient training samples for some classes in TREC50. The second TREC50 row uses TDD with 5 GD steps with 4 images per class. Experiments with random initializations have their results listed in the form [mean $\pm$ standard deviation] and are based on the resulting performance of 200 randomly initialized networks.}
\label{tab:txt_results}
\begin{tabular}{lllllllll}
        \hline
        & \multicolumn{2}{c}{TDD accuracy} & \multicolumn{4}{c}{Used as training data in 10 GD steps} & \multicolumn{2}{c}{Used in K-NN} \\
        & Fixed & Random &  Rand. real & Optim. real & $k$-means & Avg. real & Rand. real & k-means\\
        \hline
        IMDB & \textbf{75.0} & \textbf{73.4 $\pm$ 3.3} & 49.7 $\pm$ 0.9 & 49.9 $\pm$ 0.8 & 49.9 $\pm$ 0.6 & 50.0 $\pm$ 0.1 &  50.0 $\pm$ 0.1  & 50.0 $\pm$ 0.0\\
        SST5 & \textbf{37.5} & \textbf{36.3 $\pm$ 1.5} & 21.2 $\pm$ 4.9 & 24.6 $\pm$ 2.6 & 19.6 $\pm$ 4.5 & 21.3 $\pm$ 4.1 & 23.1 $\pm$ 0.0 & 20.9 $\pm$ 2.1\\
        TREC6 & \textbf{79.2} & \textbf{77.3 $\pm$ 2.9} & 37.5 $\pm$ 10.1 & 44.6 $\pm$ 7.5 & 34.4 $\pm$ 13.0 & 28.0 $\pm$ 9.5 & 31.5 $\pm$ 9.9 & 50.5 $\pm$ 6.8\\
        TREC50 & \textbf{57.6} & 11.0 $\pm$ 0.0  & 8.2 $\pm$ 6.0 & 9.9 $\pm$ 6.6 & 14.7 $\pm$ 5.5 & 12.5 $\pm$ 6.4 &  15.4 $\pm$ 5.1 & \textbf{45.1 $\pm$ 6.6}\\
        TREC50$^2$ & 67.4 & 42.1 $\pm$ 2.1 &&&&&&\\
        \hline
\end{tabular}

\end{sidewaystable}

\subsection{Text Data}
As mentioned above, TDD does not work in the space of the original raw data, but rather produces synthetic samples from the embedding space. Because each distilled `sentence' is actually a matrix, the embedding layer is applied only to real sentences from the training data and not the matrices coming from the distilled data. For text experiments, we use the IMDB sentiment analysis dataset, the Stanford Sentiment Treebank 5-class task (SST5) \citep{sst5}, and the Text Retrieval Conference question classification tasks with 6 (TREC6) and 50 (TREC50) classes \citep{trec}. The text experiments are performed with three different networks: a fairly shallow but wide CNN (TextConvNet), a bi-directional RNN (Bi-RNN) \citep{schuster1997bidirectional}, and a bi-directional Long Short-Term Memory network (Bi-LSTM) \citep{hochreiter1997long}. These models do not all achieve the same accuracies on the text datasets we use in distillation experiments, so the models' original accuracies are detailed in Table~\ref{tab:CNNaccs}. The distillation ratios appearing in this section are calculated based on these accuracies. The exact architectures of these models are detailed in the online appendix. 


\begin{table}
\caption{Model accuracies when trained on full text datasets.}
\label{tab:CNNaccs}       
\centering
\begin{tabular}{llcc}
\hline\noalign{\smallskip}
Model & Dataset & \# of Classes & Accuracy  \\
\noalign{\smallskip}\hline\noalign{\smallskip}
TextConvNet & IMDB & 2 & 87.1\%\\
Bi-RNN & SST5 & 5 & 41.0\% \\
Bi-LSTM & TREC6 & 6 & 89.4\% \\
TextConvNet & TREC50 & 50 & 84.4\% \\
\noalign{\smallskip}\hline
\end{tabular}
\end{table}
\begin{table}
\caption{Distillation ratios for text datasets and their associated neural networks. Experiments with random initializations have their results listed in the form [mean $\pm$ standard deviation] and are based on the resulting performance of 200 randomly initialized networks.}
\label{tab:rnn_results}
\centering
\begin{tabular}{lllll}
        \hline
        & & Number of &\multicolumn{2}{c}{Distillation Ratio ($r_M$)} \\
        Model & Dataset & Distilled Sentences (M) & Fixed & Random  \\
        &&&&(Mean $\pm$ SD)\\
        \hline
        TextConvNet & IMDB & 2 & 89.9 & 80.0 $\pm$ 6.3\\
        TextConvNet & IMDB & 20 & 91.5 & 85.2 $\pm$ 3.2\\
        Bi-RNN & SST5 & 5 & 87.7 & 57.0 $\pm$ 5.7\\
        Bi-RNN & SST5 & 100 & 89.8 & 66.8 $\pm$ 5.4\\
        Bi-LSTM & TREC6 & 6 & 97.8 & 69.3 $\pm$ 9.8\\
        Bi-LSTM & TREC6 & 120 &98.2 & 78.9 $\pm$6.3\\
        TextConvNet & TREC50 & 500 &57.6 & 11.0 $\pm$ 0.0\\
        TextConvNet & TREC50 &1000& 67.4 & 42.1 $\pm$ 2.1\\
        \hline
\end{tabular}
\end{table}
\clearpage
\paragraph{Baselines.}
We consider the same six baselines as in the image case but modify them slightly so that they work with text data. 
\begin{itemize}
    \item \textbf{Random real sentences:} We randomly sample the same number of real sentences per class, pad/truncate them, and look up their embeddings. These sentences are used for two baselines: training neural networks and training K-Nearest Neighbors classifiers.
    \item \textbf{Optimized real sentences:} We sample and pre-process different sets of random real sentences as above, but now we choose the 20\% of the sets that have the best performance. These sentences are used for one baseline: training neural networks.
    \item \textbf{$k$-means:} First, we pre-process the sentences. Then, we use $k$-means to learn clusters for each class, and use the resulting centroids to train. These sentences are used for two baselines: training neural networks and training K-Nearest Neighbors classifiers.
    \item \textbf{Average real sentences:} First, we pre-process the sentences. Then, we compute the average embedded matrix for each class and use it for training. These sentences are used for one baseline: training neural networks.
\end{itemize}
Each of these baseline methods produces a small set of sentences, or sentence embeddings, that can be used to train models. All four of the baseline methods are used to train and test the TextConvNet on each of the text datasets. Additionally, two of the baseline methods are used to also train K-Nearest Neighbor classifiers to compare performance against neural networks. The baseline results are shown in Table~\ref{tab:txt_results}.

\begin{figure}
    \centering
    \includegraphics[width=\textwidth, trim={2.8cm 2.5cm 1.9cm 0.9cm}, clip]{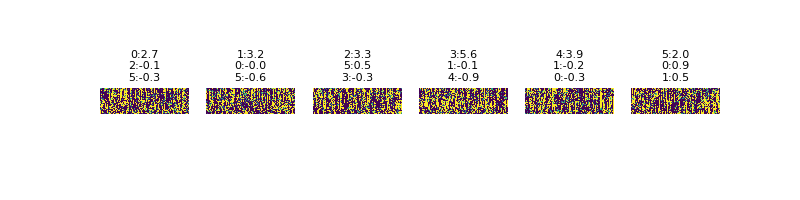}
    \caption{TDD can learn 6 distilled sentences of length 30 that train networks with fixed initializations from 12.6\% to 87.4\% ($r_6=97.8$). Each image corresponds to a distilled embedding and is labeled with its top 3 classes and their associated logits. }
    \label{fig:trec_known}
\end{figure}

\begin{figure}
    \centering
    \includegraphics[scale=0.8, trim={0cm 0.6cm 0cm 0.1cm}, clip]{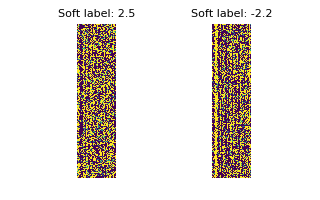}
    \caption{TDD can learn 2 distilled sentences of length 400 that train networks with random initializations from 50.0\% to $69.6\% \pm 5.5\%$ ($r_2=79.96$). Each image corresponds to a distilled embedding and is labeled with its soft label value. The soft label is a scalar as this is a binary classification task. }
    \label{fig:imdb_unknown}
\end{figure}

\paragraph{Fixed initialization.}
When the network initialization is fixed between the distillation and training phases, synthetic text produced by text dataset distillation also results in high model accuracies. For example, TDD can produce 2 distilled sentences that train the TextConvNet up to a distillation ratio of 89.88\% on the IMDB dataset. Even for far more difficult language tasks, TDD still has impressive results but with larger distilled datasets. For example, for the 50-class TREC50 task, it can produce 1000 distilled sentences that train the TextConvNet to a distillation ratio of 79.86\%. Some examples of TDD performance are detailed in Table~\ref{tab:txt_results} and Table~\ref{tab:rnn_results}. The distilled text embeddings from the six-sentence Trec6 experiments are visualized in Figure~\ref{fig:trec_known}. However, since these distilled text embeddings are still in the GloVe embedding space, it may be difficult to interpret them visually. We provide a more natural method for analyzing distilled sentences by using nearest-word decoding to reverse the GloVe embedding. We find the nearest word to each distilled vector based on Euclidean distances. The result of this decoding is an approximation of the distilled sentence in the original text space. We list the decoded distilled sentences corresponding to the matrices from Figure~\ref{fig:trec_known} in Table~\ref{tab:trec_dist_sent}, along with their respective label distributions. These sentences can contain any tokens found in the TREC6 dataset, including punctuation, numbers, abbreviations, etc. The sentences do not have much overlap. This is consistent with the distilled labels which suggest that each sentence corresponds strongly to a different class. It appears that the TDD algorithm encourages the separation of classes, at least when there are enough distilled samples to have one or more per class. Additional results and visualizations for TDD with fixed initialization can be found in the online appendix. 
\clearpage
\begin{table}[]
    \centering
    \begin{tabular}{|p{85mm}|c|c|c|c|c|c|}
    \hline
          & \multicolumn{6}{c|}{\textbf{Label Class}} \\
          \textbf{Distilled Sentence} &\textbf{0}&\textbf{1}&\textbf{2}&\textbf{3}&\textbf{4}&\textbf{5}\\
          \hline
          allan milk banned yellow planted successfully introduced bombay 1936 grass mines iron delhi 1942 male heir throne oath clouds 7th occur millennium smoking flows truth powder judiciary pact slim profit & \textbf{2.72}&-0.48&-0.07&-0.62&-0.53&-0.27\\
          \hline
         whom engineer grandfather joan officer entered victoria 1940s taxi romania motorcycle italian businessman photographer powerful driving u brilliant affect princess 1940s enemies conflicts southwestern retired cola appearances super dow consumption&-0.05&\textbf{3.21}&-1.15&-0.79&-0.71&-0.64\\
         \hline
          necessarily factors pronounced pronounced define bow destroying belonged balls 1923 storms buildings 1925 victorian sank dragged reputation sailed nn occurs darkness blockade residence traveled banner chef ruth rick lion psychology&-0.67&-0.66&\textbf{3.28}&-0.27&-0.77&0.47\\
          \hline
         accommodate accommodate peak 2.5 adults thin teenagers hike aged nurse policeman admit aged median philippines define baghdad libya ambassador admit baseman burma inning bills trillion donor fined visited stationed clean &-0.98&-0.14&-1.12&\textbf{5.57}&-0.85&-1.85\\
         \hline
          suburb ports adjacent mountains nearest compare hilton volcano igor nebraska correspondent 1926 suburb sailed hampshire hampshire gathering lesson proposition metric copy carroll sacred moral lottery whatever fix o completed ultimate&-0.29&-0.22&-0.36&-1.01&\textbf{3.86}&-0.44\\
          \hline
         advertising racism excuse d nancy solved continuing congo diameter oxygen accommodate provider commercials spread pregnancy mideast ghana attraction volleyball zones kills partner serves serves congressman advisory displays ranges profit evil &0.94&0.53&-0.85&0.08&-0.83&\textbf{1.99}\\
         \hline
    \end{tabular}
    \caption{Nearest sentence decodings corresponding to the distilled embeddings in Figure~\ref{fig:trec_known}. Each sentence is accompanied by the logits associated with each value of its distilled label. The top class for each distilled sentence is bolded.}
    \label{tab:trec_dist_sent}
\end{table}
\clearpage
\paragraph{Random initialization.}
When using random initialization the performance decrease for TDD is similar to that for SLDD. TDD can produce two distilled sentences that train the TextConvNet with random initialization up to a distillation ratio of 79.96\% on IMDB, or 20 distilled sentences that train it up to a distillation ratio of 85.22\%. This is only slightly lower performance than in the fixed initialization case. However, there is a larger difference in performance between fixed and random initializations for the recurrent networks. For TREC6, TDD can produce six distilled sentences that train a randomly initialized Bi-LSTM to a distillation ratio of 69.33\%, or 120 distilled sentences that train it to a distillation ratio of 78.87\%. All the mean and standard deviation results for random initializations in Table~\ref{tab:txt_results} and Table~\ref{tab:rnn_results} are derived by testing with 200 randomly initialized networks. The distilled text embeddings from the IMDB experiment with two distilled sentences are visualized in Figure~\ref{fig:imdb_unknown}. We list the decoded distilled sentences corresponding to these matrices in Table~\ref{tab:imdb_dist_sent}. These sentences can contain any tokens found in the IMDB dataset, including punctuation, numbers, abbreviations, etc. Since this is a binary sentiment classification task, each label is a scalar. If a probability is needed, a sigmoid function can be applied to the scalar soft labels. In this case, the distillation algorithm appears to have produced one sentence with a positive associated sentiment, and one with a negative sentiment. Curiously, the model appears to have overcome the challenge of having to describe these long sentences with a single scalar by using duplication. For example, in the second sentence, corresponding to the negative label, negative words like `dump', `stupid', and `shoddy' are all repeated several times. In such a way, the algorithm is likely assigning lower sentiment scores to these words than other ones, while using only a single label. Additional results for TDD with random initialization can be found in the online appendix. 

\begin{table}[]
    \centering
    \begin{tabular}{|p{145mm}|c|}
    \hline
          \textbf{Distilled Sentence} &\textbf{Label}\\
          \hline
          editor panoramic brewing swat regency medley arts fleet attained hilary novelist rugged medal who abbot has sweden new ensemble member understands alba archaeologist operatic intercontinental martian marshal paste smooth titular english-language adaptation songwriter historian enlightenment royal gaelic ceo author macarthur skipper honored excellence distance most endings collaborations his mythological polanski mayer choreographer grisham eminent brooke sympathies modelling vitality dictionary dedication farewell enjoys energetic jordan equality lectures sophia elijah maureen novelist sanjay zeta blues colors talent dylan dominic anne 2003 who pavarotti dedication illustration mary exhibited scenic award colonial filmography previews dialect scenery steely representative adventures opens favourites macarthur economic erected mabel biography def waterfront awards tenor artists coin choreographer art securing anna earned helen emmy vibrant repertoire witty prestigious marquis classic splendor kindness lyrical architecture climbed princess quotes ava touring eponymous johnson strengths title county acclaimed courage revolt prolific classic honors rocker unforgettable ageing founding vigorous renowned deepest operatic female intertwined impressionist poignant vanilla impassioned esp isle camaraderie romania regal casablanca caribbean friendship friendship todd meadows hosted billie regina intimate poetic digital warmth savvy great symphony twilight thrills meditation ebert emperor snail remarkable rhys classics talent kristin patricia kim ballads climber jordan advancing phil i. robbie beatrice brighter pacifist hebrew & 2.53\\
          \hline
         que yeah dump misled speculate bait substandard uncovered lump corpses 911 punched whopping discarded ref dollar were dough sided brink unconscious tomatoes locking trash burying punched diversion grenade overboard cashier wards agreeing brent prematurely knife randomly stupid buster wipe virus pap waste inflated loan patient peg eliminates nudity worms ?? rotten shoddy strangled substandard boil whopping tunnels steep unjust dummy satisfactory mistakenly adopt tightened bloated hacked misled dump 3-4 untrue contaminated bureaucracy waste chopping bait gram rotting intentional washed unhealthy liar corpses hospitals ?? filthy lifts freeze cabin contaminated shutting rents discrimination frying stomach toxic piles riddled negligible shattering fewer radioactive melt censors gram spilling housing saif subway sabotage booby nuke bb wasteland nuisance knock false useless gonna budget useless grenades cigarettes financed mandatory mine parked bb !! hangover riddled hijack shut cheaply unused sim mole trailers collide misunderstanding lump projected litter fx worst tamil rotten sewer substandard financing leak ?? gonna any claiming tails giancarlo explode frustrating flooding slit nixon prematurely pay leftover chimney outs shotgun stupid surplus minimum shutting punched radioactive stupid worthless worst shoddy nudity rotting workable worthless doses hanged rub be busted stupid faulty lame gunshot grenade dump trash wrist alastair hurry suffice detect rigged transformers unrated > deplorable scrape hacked & -2.21\\
         \hline
    \end{tabular}
    \caption{Nearest sentence decodings corresponding to the distilled embeddings in Figure~\ref{fig:imdb_unknown}. Each sentence is accompanied by its associated soft label. Only the first 200 (out of 400) words are shown for each sentence. }
    \label{tab:imdb_dist_sent}
\end{table}


\section{Conclusion}\label{con}
 By introducing learnable distilled labels we have increased distillation accuracy across multiple datasets by up to 6\%. By enabling text distillation, we have also greatly increased the types of datasets and architectures with which distillation can be used.
 
 However, even with SLDD and TDD, there are still some limitations to dataset distillation. The network initializations used for both SLDD and TDD all come from the same distribution, and no testing has yet been done on whether a single distilled dataset can be used to train networks with different architectures. Further investigations are needed to determine more precisely how well dataset distillation can be generalized to work with more variation in initializations, and even across networks with different architectures.  
 
 Interestingly, the initialization of distilled labels appears to affect the performance of dataset distillation. Initializing the distilled labels with `hard' label values leads to better performance than with random initialization, possibly because it encourages class separation earlier on in the distillation process. However, it is not immediately clear whether it is better to separate similar classes (e.g. `3' and `8' in MNIST), thereby increasing the network's ability to discern between them, or to instead keep those classes together, thereby allowing soft-label information to be shared between them. It may be interesting to explore the dynamics of the distillation process when using a variety of label initialization methods. 
 
 We have shown dataset distillation works with CNNs, bi-directional RNNs, and LSTMs. There is nothing in the dataset distillation algorithm that would limit it to these network types. As long as a network has a twice-differentiable loss function and the gradient can be back-propagated all the way to the inputs, then that network is compatible with dataset distillation. 
 
 Another promising direction is to use distilled datasets for speeding up Neural Architecture Search and other very compute-intensive meta-algorithms. If distilled datasets are a good proxy for performance evaluation, they can reduce search times by multiple orders of magnitude. In general, dataset distillation is an exciting new branch of knowledge distillation; improvements may help us not only better understand our datasets but also enable several applications related to efficient machine learning.



\acks{We would like to thank Dr. Sebastian Fischmeister for providing us with the computational resources that enabled us to perform many of the experiments found in this work.  }


\newpage








\bibliography{idd}

\end{document}